%% file: report.tex
\documentclass[12pt,a4paper]{article}

\usepackage{amssymb}
\usepackage{amsfonts}
\usepackage{amsmath}
\usepackage{graphicx}
\usepackage{subcaption}
\usepackage{multirow}
\usepackage{algorithm}
\usepackage{algpseudocode}
\usepackage{bm}
\usepackage{varwidth}
\usepackage{placeins}

 \usepackage[disable]{todonotes} 

\algtext*{EndFor}
\algtext*{EndIf}

 \newcommand{\comment}[1]{}  

\newcommand{\qed}{\rule{1.5mm}{2mm}\vspace{0.1in}}

\newcommand{\ignore}[1]{}

\usepackage{xcolor}

\title{Behavior-Based Machine-Learning: \\ A Hybrid Approach \\ for Predicting Human Decision Making}

\author{
Gali Noti\thanks{Rachel \& Selim Benin School of Computer Science \& Engineering and Federmann Center for the Study of Rationality, The Hebrew University of Jerusalem, Israel.}
\and
Effi Levi\thanks{Rachel \& Selim Benin School of Computer Science \& Engineering, The Hebrew University of Jerusalem, Israel.}
\and
Yoav Kolumbus\thanks{Racah Institute of Physics, The Hebrew University of Jerusalem, Israel.} 
\and 
Amit Daniely\thanks{Rachel \& Selim Benin School of Computer Science \& Engineering, The Hebrew University of Jerusalem, Israel; and Google.}
}

\date{}

\begin{document}
\maketitle

\begin{abstract}
A large body of work in behavioral fields attempts to develop models that describe the way people, as opposed to rational agents, make decisions. 
A recent Choice Prediction Competition (2015) challenged researchers to suggest a model that captures 14 classic choice biases and can predict human decisions under risk and ambiguity. 
The competition focused on simple decision problems, in which human subjects were asked to repeatedly choose between two gamble options. 

In this paper we present our approach for predicting human decision behavior: 
we suggest to use machine learning algorithms with features that are based on well-established behavioral theories. 
The basic idea is that these psychological features are essential for the representation of the data and are important for the success of the learning process.
We implement a vanilla model in which we train SVM models using behavioral features 
that rely on the psychological properties underlying the competition baseline model. 
We show that this basic model captures the 14 choice biases and outperforms all the other learning-based models in the competition. 
The preliminary results suggest that such hybrid models 
can significantly improve the prediction of human decision making, and are a promising direction for future research. 
\end{abstract}

\section{Introduction}
\input{intro}


\section{The Hybrid Model}
\input{model}

\section{Results and Conclusion} \label{sec_results}
\input{conclusion}

\bibliographystyle{abbrv}
\bibliography{report}

\listoftodos

\end{document}

%% file: intro.tex
It is well known that the classic economic theory, which considers perfectly rational decision makers, generally fails to predict human decisions. 
A large body of work in psychology and behavioral economics attempts to capture the way humans make decisions. These attempts typically address behavior patterns that consistently emerge in a specific setting, but often these patterns are different, or even contradicting, when the setting is changed.  
For instance, in risky choice problems, when the decision is based solely on the description of the problem people tend to overweight rare events \cite{prospect_theory}, while in decisions based on experience they tend to underweight rare events (see review of the experience-description gap in \cite{rare_events}). 

Recently, researchers were challenged in the Choice Prediction Competition (CPC) 2015 \cite{cpc2015} to propose a model that will be general enough to predict human decisions for a wide space of problems. 
The competition focused on 14 well-studied choice anomalies, 
which included the four biases captured by prospect theory \cite{prospect_theory} (including overweighting of rare events), and additional 10 biases that were documented in studies of decisions with and without feedback (including underweighting of rare events in repeated settings with feedback).
The organizers, Erev et al. (2015), defined a problem space for which they showed that these anomalies emerge (in a series of experiments). 
The challenge for the competitors was to develop a model that would best
predict the decisions made by humans in any problem in this problem space. 
For the full description of the competition see \cite{cpc2015} and the competition website. Next we briefly describe the main details.

\noindent {\bf The Problem Space: } \label{sec_space}
The organizers defined an 11-dimensional problem space of choice tasks that is wide enough to give rise to the 14 choice anomalies discussed above. 
Denote the problem space by $PS = LA \times HA \times pHA \times LB \times HB \times pHB \times LotNum \times LotShape \times Corr \times Amb \times FB$. 
Each problem $x \in PS$ is a choice between two gamble options A and B. Option A provides the high outcome $HA$ with probability $pHA$, and the low outcome $LA$ otherwise. Option B provides with probability $pHB$, an outcome of a lottery with an expected value $HB$, or $LB$ otherwise. $LotNum$ defines the number of possible outcomes in the lottery and $LotShape$ defines whether the lottery's distribution is symmetric, right-skewed, or left-skewed around its mean (or is undefined if $LotNum=1$). $Corr$ determines the sign of the correlation between the payoffs of the two options.

The 10th dimension, Ambiguity ($Amb$), defines the precision of the initial information the decision maker receives regarding the probabilities of the outcomes in Option B. The competition focused on the two extreme cases: $Amb=0$ implies complete information (i.e., no ambiguity), and $Amb=1$ implies that the decision maker does not receive any initial information regarding option B's probabilities. The 11th dimension is the feedback to the decision maker and was studied within a problem: subjects repeatedly faced each problem, first without any feedback and then with complete feedback.

\noindent {\bf Data: } 
Prior to the competition announcement, the organizers performed two preliminary experiments. 
In each experiment, human participants were presented with a set of decision problems drawn from the problem space. 
Participants faced each problem for 25 trials. In the first five trials they did not receive any feedback, while 
following each of the remaining 20 trials they received 
full feedback of the obtained and forgone outcomes. The 25 trials were divided to five blocks, each consisting of five consecutive trials, and the average portion of option B choosers (the {\em B-rate}) was recorded for each block. 
In the first experiment, 30 hand-tailored problems in the problem space were used for replicating the emergence of the 14 choice anomalies mentioned above. The second experiment used 60 problems that were randomly generated from the problem space, to show the robustness of the anomalies.

\noindent {\bf The Challenge: } 
Given the parameters and B-rates for these 90 decision problems (the {\em estimation set}), the competitors were asked to provide a model (in the form of a program), which receives a problem represented by the first 10 parameters described above\footnote{The 11th parameter is tested within a problem.} and outputs five B-rates, one for each block. The model must capture all 14 choice anomalies (represented by a published script containing 14 tests which the model should pass), and should be easy and straight-forward to implement (from a description of at most 1500 words).

To evaluate the models, the organizers ran a third experiment using 60 new randomly generated decision problems. The results of this experiment were used for evaluating the competitors' models (the {\em test set}), 
using the Mean Squared Deviation (MSD) score. For a set of observations $OBS_j$ and a set of predictions $B_j$ for problems $j = 1..N$:

\begin{equation}
	MSD(OBS, B) = 100\cdot\frac{1}{N}\sum_j{(OBS_j-B_j)^2}
\end{equation}

The final evaluation metric is the MSD averaged over all five blocks.

\noindent {\bf Our Approach: } 
The common approach towards solving this problem in behavioral fields is to derive 
 patterns from behavior observed in experiments (in some setting), 
and then define a model which employs these 
patterns to make a prediction, all generally based on well-established literature 
(see, for example, the competition baseline model \cite{cpc2015}). 
These approaches put the emphasis on characterizing or describing the 
observed behavior, 
but generally use simple, hand-tuned methods to make the actual prediction.
 
We choose to take a different approach towards modeling human decision making. We use the behavioral theory based patterns to compute a feature representation for the decision problem, to be used as input to powerful machine learning algorithms. In this prospect, the computational field of machine learning has much to offer, from sophisticated learning algorithms to efficient ways for feature selection and reduction, as well as reliable validation techniques. 

In this work we implement a vanilla version of a hybrid model that combines Support Vector Machines (SVM) models with behavioral-based features and adjustments. The behavioral features are computed based on an {\em effective} (i.e., biased) form of the target problem.
Our hybrid model was ranked lower than most of the variants of the baseline model, but outperformed all the learning-based models in the competition,  
demonstrating that these psychological features are important for the success of the learning process.

In the next section we present the details of our hybrid model and in Section \ref{sec_results} we present the competition results and conclude with future research directions. 

%
%
%
%
%
%
%

%% file: model.tex

\subsection{Feature representation} \label{sec_ftrs}
Each sample representing a problem $p$ is given by 10 parameters: $LA, HA, pHA,$\\
$LB, HB, pHB, LotNum, LotShape, Corr, Amb$ (the first 10 coordinates of the problem space $PS$).
For each block $i$ of consecutive five choice trials, we compute a feature representation $f^i_p$ that is consisted of (24+i-1) features, as follows. 

First, we generate a problem $p^i_{eff}$ by converting $p$'s probabilities -- $pHa$ and $pHb$ -- to their $effective form$ at block $i$. 
	Specifically, set $pHa = effective(pHa,i)$ and $pHb = effective(pHb,i)$, by:
	$$ effective(p,i) = \begin{cases}
    \frac{p}{i} & \text{if } p \leq 0.02 \\
    p + (1-\frac{1}{i}) \cdot (1-p) & \text{if } p \geq 0.98 \\
    p & \text{otherwise } 
  \end{cases} $$ 
	That is, the effective form of a problem reflects the tendency to underweight rare probabilities as $i$ increases. 
	
Next, we compute a list of 16 behavior-based features from the parameters of $p^i_{eff}$. These features were designed to capture salient properties of the problem, which play a role when people face the problem of choosing between options A and B, in our setting of small decisions under risk and ambiguity. Most of these features are derived from the baseline model provided by \cite{cpc2015}. The features include properties of a single option as well as properties that refer to the problem as a whole.

The features make use of the full distributions of options A and B. Let us annotate these distributions by $DistA=(o^a_1,p^a_1;o^a_2,p^a_2;...;o^a_{s_a},p^a_{s_a})$ and $DistB=(o^b_1,p^b_1;o^b_2,p^b_2;...;o^b_{s_b},p^b_{s_b})$, i.e., the full outcomes and probabilities, where the outcomes $o^*_j$ appear in ascending order. We proceed to describe all the features.
	\begin{enumerate}
		
		\item $IsGain$: equals 1 if the portion of non-negative outcomes is greater than $0.65$ and at least one outcome is strictly positive, or 0 otherwise. 
	
	\item $IsLoss$: equals 1 if the portion of non-positive outcomes is greater than $0.65$ and at least one outcome is strictly negative, or 0 
	otherwise. 
	
	\item $EvA$: The expected value of option A. Specifically, $EvA = \sum^{s_a}_{j=1} o^a_j \cdot p^a_j$.
	
	\item $BevB$: The best estimation of the expected value of option B, as follows:
	$$BevB =  \begin{cases}
    0.48 \cdot\sum^{s_b}_{j=1} \frac{1}{s_b}o^b_j + 0.48\cdot EvA + 0.04 \cdot o^b_1 & \text{if } i=1 \text{ and } Amb=1\\
		\sum^{s_b}_{j=1} o^b_j \cdot p^b_j & \text{otherwise } 
  \end{cases} $$
	For ambiguous problems in the first round, $BevB$ is a pessimistic estimation of the expectation assuming pHB is uniform, and biased towards option A's expected value. Otherwise, it is the actual expected value of option B.
	
	\item $BevA\_pt$: The expected value of option A, using prospect theory valuation and weighting functions. Specifically, 
	$$BevA\_pt = V(o^a_1)\pi(p^a_1) + ... + V(o^a_{s_a})\pi(p^a_{s_a})$$
	where 
	$$V(o_i) = \begin{cases}
    o^\alpha_i & \text{if } o_i \geq 0\\
		-\lambda|o_i|^\beta & \text{otherwise } 
  \end{cases} $$
	and
	$$\pi(p_i) = \begin{cases}
    \frac{p^\gamma_i}{(p^\gamma_i+(1-p_i)^\gamma)^{\frac{1}{\gamma}}} & \text{if } o_i \geq 0\\
		\frac{p^\delta_i}{(p^\delta_i+(1-p_i)^\delta)^{\frac{1}{\delta}}} & \text{otherwise } 
  \end{cases} $$
	we used $\alpha=0.77,\beta= 0.9,\gamma=0.79,\delta=0.87, \lambda=1.0023$.
	
	\item $BevB\_pt$: Same as $BevA\_pt$, computed for DistB with one small change: for ambiguous problems in the first round, we take DistB to be
	$$p^b_1 = \frac{bevB - \frac{1}{s_b}\sum_{j=2}^{s_b}{o^b_j}}{o^b_1 - \frac{1}{s_b}\sum_{j=2}^{s_b}{o^b_j}}$$
	and
	$$p^b_2 = p^b_3 = ... = p^b_{s_b} = \frac{1-p^b_1}{s_b-1}$$
	
	\item $SignMax$: The sign of the maximal outcome. 
	
	\item $RatioMin$: 
	$$RatioMin = \begin{cases}
		\frac{min\{|o^a_1|,|o^b_1|\}}{max\{|o^a_1|,|o^b_1|\}} & \text{if } sign(o^a_1)=sign(o^b_1)\\
		0 & \text{otherwise } 
	\end{cases} $$

	\item $BevBminA$: The difference between the expected value of option B and option A. That is, 
	$$BevBminA = BevB - EvA$$
	
	\item $BevBminA\_pt$: The difference between the prospect-theory expected value of option B and option A. That is,
	$$BevBminA\_pt = BevB\_pt - EvA\_pt$$
	
	\item $BevBminA\_unif$: The difference between the expected value of option B and option A, assuming the outcomes are distributed uniformly in both options. 
	Specifically, 
	$$BevBminA\_unif = \sum^{s_b}_{j=1}{\frac{1}{s_b} o^b_j} - \sum^{s_a}_{j=1}{\frac{1}{s_a} o^a_j}$$
	
	\item $BevBminA\_sign$: The difference between the expected value of option B and of option A, assuming that positive outcomes are replaced by $R$ and negative outcomes are replace by $-R$, where $R$ is the payoff range. That is, 
	$$BevBminA\_sign = \sum^{s_b}_{j=1}{R\cdot sgn(o^b_j)\cdot p^b_j} - \sum^{s_a}_{j=1}{R\cdot sgn(o^a_j)\cdot p^a_j}$$
	where
	$R = max\{o^a_1,...,o^a_{s_a},o^b_1,...,o^b_{s_b}\} - min\{o^a_1,...,o^a_{s_a},o^b_1,...,o^b_{s_b}\}$. 
	
	\item $BisBetter$: The probability that B will give a better outcome, computed as:
	$$ BisBetter = \sum_{i=1}^{s_b}\left({p_b^i \cdot \sum_{j : o^a_j < o^b_j}{p^a_j}}\right) $$
	
	\item $SCPT$: The decision made by the Stochastic Cumulative Prospect Theory model described in \cite{scpt}.
	
	\item $VarBminA$: The difference in variance between option B and option A.
	
	\item $EntBestMinA$: The difference in entropy between option B and option A, where the entropy for a distribution $p={p_1, ..., p_s}$ is computed as 
	$$entropy(p) = -\sum_{i=1}^{s}{p_i \log p_i}$$

	\end{enumerate}

	Finally, the feature representation $f^i_p$ for a problem $p$ in block $i$ is a vector which is obtained by 
	
	\begin{itemize}
	\item Setting the first 8 entries to $p$'s values: $(Ha,pHa,La,Hb,pHb,Lb,Amb,Corr)$.
	\item Setting entries 9-24 to the features listed above based on $p^i_{eff}$.
	\item Setting the next $i-1$ entries to the B-rates predicted in the $i-1$ previous blocks $(B_1,...,B_{i-1})$ (or to the B-rates observed in the estimation experiments\footnote{Experiment 1 and 2, data in Tables 2 and 3, in \cite{cpc2015}, respectively.} when computed for the training set).
	\end{itemize}
		
	We use the standardized vector, where each of the first 24 entries is standardized to have zero mean and unit variance based on randomly generated 100,000 problems (from the problem space described in Section \ref{sec_space}), and the $B_i$s are standardized based on the 90 estimation problems. 


\subsection{Learning Algorithm}
For learning the problem we chose to use Support Vector Machines (SVM) \cite{svm}, which is a versatile and powerful learning method.
We experimented with various choices of kernels, and the best results were obtained with a polynomial kernel of the third degree, using N-fold (for $N=10$) cross validation over the 90 estimation problems.
For each block $i \in \{1,...,5\}$ we learned a separate SVM model over the feature representation at block $i$ as described above.

\subsection{Prediction}
Given a target problem $p$ represented by a feature vector $x$, 
the hybrid model predicts the B-rates for $x$ sequentially: first predicting $B_1$, the average B-rates in the first block of ``from description'' trials, 
then predicting every later $B_i$ at block $i \in \{2,...,5\}$ of ``from feedback'' trials, relying on the predictions so far ($B_1,...,B_{i-1}$). The prediction for each block $i$ is done using the SVM model learned for block $i$ followed by adjustments based on the problem properties. 

Specifically, for each block $i\in \{1,...,5\}$, the hybrid model predicts B-rates $B_i$ by carrying the following procedure:
\begin{enumerate}
	
	\item Set $z$ to be the standardized feature representation of the target problem $x$.
	
	\item Apply the SVM model at block $i$ to $z$ to obtain $B_i$.
	
	\item Adjust $B_i$ in the following two cases, taking into account the effective form of the problem $p^i_{eff}$, and the features $BevB$, $EvA$ and $SCPT$ of $x_{eff}$ (see Section \ref{sec_ftrs}). 
	\begin{enumerate}
	\item If $p^i_{eff}$ is trivial, i.e., if one of the options dominates the other, then average $B_i$ with the trivial choice. 
	\item If $i = 1$, and at least one of the conditions $|B_i-0.5|<0.025$ or $BevB=EvA$ holds, then set $B_i=0.7*B_i+0.3*SCPT$ (where SCPT is the feature calculated in Section~\ref{sec_ftrs}). 
	
	We therefore push the prediction in the direction suggested by Stochastic Cumulative Prospect Theory (SCPT) when the B-rate prediction in the first block is not clear enough.
	\end{enumerate}

\end{enumerate}

%% file: conclusion.tex
\subsection{Results}
The hybrid model passed the tests provided by \cite{cpc2015}, indicating success in capturing all 14 choice anomalies.
The MSD score on the estimation set (the 90 problems initially supplied to the competitors) was 0.74 compared to 0.71 achieved by the baseline model, with an average of 0.94 and standard deviation 0.74 over the 25 models submitted to the competition. 
The MSD score on the test set (60 new problems) was 1.24 compared to 0.98 achieved by the baseline model, with an average of 1.47 and standard deviation 0.89. 

Out of the 25 models submitted to the competition, 14 were variants of the baseline model and the remaining were based on statistics, machine learning or other methods. 
Our model ranked 14th, where ranks 1-13 were all taken by variants of the baseline model.

\subsection{Conclusion}

Our hybrid model outperformed all the models which were not variants of the baseline model. In particular, it performed better than all of the learning-based models that did not make use of behavioral theories. These results suggest that learning methods alone are not sufficient for predicting human decisions, and that representing and manipulating the data based on behavioral knowledge is important for the success of these methods, supporting our approach of integrating behavioral features and adjustments with the powerful machine learning tools to improve their performance.

However, the fact that the hybrid model was ranked lower than the baseline model and its variants, indicates that this integration should be further studied and improved. We believe that a first major improvement to the preliminary results shown in this paper can be obtained by further experimenting with the full range of parameters available for SVM, as well as trying other machine learning methods.
A second significant improvement may be obtained by learning from a larger training set. Running large scale experiments is becoming more applicable with the growing popularity of online labor markets such as Amazon Mechanical Turk. Obviously, a further study of such hybrid models for predicting human decision making is called for.